\documentclass[letterpaper]{article} 
\usepackage{aaai2026}  
\usepackage{times}  
\usepackage{helvet}  
\usepackage{courier}  
\usepackage[hyphens]{url}  
\usepackage{graphicx} 
\usepackage{natbib}  
\usepackage{caption} 
\usepackage{algorithm}
\usepackage{algorithmicx}
\usepackage{algpseudocode}
\floatname{algorithm}{Algorithm}

\usepackage{subcaption} 
\usepackage{tabularx}  
\usepackage{array}     

\usepackage{algorithmicx}
\usepackage{algpseudocode}

\usepackage{graphicx}
\usepackage{booktabs}
\usepackage{multirow}
\usepackage{amsmath}
\usepackage{graphicx}
\usepackage[table,xcdraw]{xcolor}

\usepackage{newfloat}
\usepackage{listings}
\DeclareCaptionStyle{ruled}{labelfont=normalfont,labelsep=colon,strut=off} 
\floatstyle{ruled}
\newfloat{listing}{tb}{lst}{}
\floatname{listing}{Listing}
\title{META-RAG: Meta-Analysis-Inspired Evidence-Re-Ranking Method for Retrieval-Augmented Generation in Evidence-Based Medicine}
\author{
    Mengzhou Sun\textsuperscript{\rm 1},
    Sendong Zhao\textsuperscript{\rm 1},
    Jianyu Chen\textsuperscript{\rm 1},
    Haochun Wang\textsuperscript{\rm 1},
    Bing Qin\textsuperscript{\rm 1}
}
\affiliations{
    \textsuperscript{\rm 1}Faculty of Computing, Harbin Institute of Technology\\

    mzsun@ir.hit.edu.cn, sdzhao@ir.hit.edu.cn

}

\usepackage{bibentry}

\begin{document}

\maketitle

\begin{abstract}
Evidence-based medicine (EBM) holds a crucial role in clinical application. Given suitable medical articles, doctors effectively reduce the incidence of misdiagnoses. Researchers find it efficient to use large language models (LLMs) techniques like RAG for EBM tasks. However, the EBM maintains stringent requirements for evidence, and RAG applications in EBM struggle to efficiently distinguish high-quality evidence. Therefore, inspired by the meta-analysis used in EBM, we provide a new method to re-rank and filter the medical evidence. This method presents multiple principles to filter the best evidence for LLMs to diagnose. We employ a combination of several EBM methods to emulate the meta-analysis, which includes reliability analysis, heterogeneity analysis, and extrapolation analysis. These processes allow the users to retrieve the best medical evidence for the LLMs. Ultimately, we evaluate these high-quality articles and show an accuracy improvement of up to 11.4\% in our experiments and results. Our method successfully enables RAG to extract higher-quality and more reliable evidence from the PubMed dataset. This work can reduce the infusion of incorrect knowledge into responses and help users receive more effective replies.
\end{abstract}

\section{Introduction}
Currently, Evidence-Based Medicine (EBM) is gradually being embraced by doctors as an essential discipline in the medical field~\cite{subbiah2023next}. Using EBM can significantly reduce the risk of misdiagnosis by referring to the retrieved medical articles. As the volume of medical evidence grows, doctors start to rely on artificial intelligence (AI) technology to assist in the practice of EBM~\cite{djulbegovic2017progress}. The key requirement from AI is to leverage all available resources, extracting and synthesizing all relevant evidence to arrive at a comprehensive conclusion~\cite{clusmann2023future}. However, due to the limitation of memory capacity, small-scale models often struggle to deal with a large amount of evidence~\cite{friedman2013natural,nadkarni2011natural}. Recently, Large Language Models (LLMs) have been presented, which are equipped with a long input restriction and exceptional comprehension ability. There have been breakthroughs in using LLMs to assist EBM.



\begin{figure}[t]
    \centering
    \includegraphics[width=1.0\linewidth]{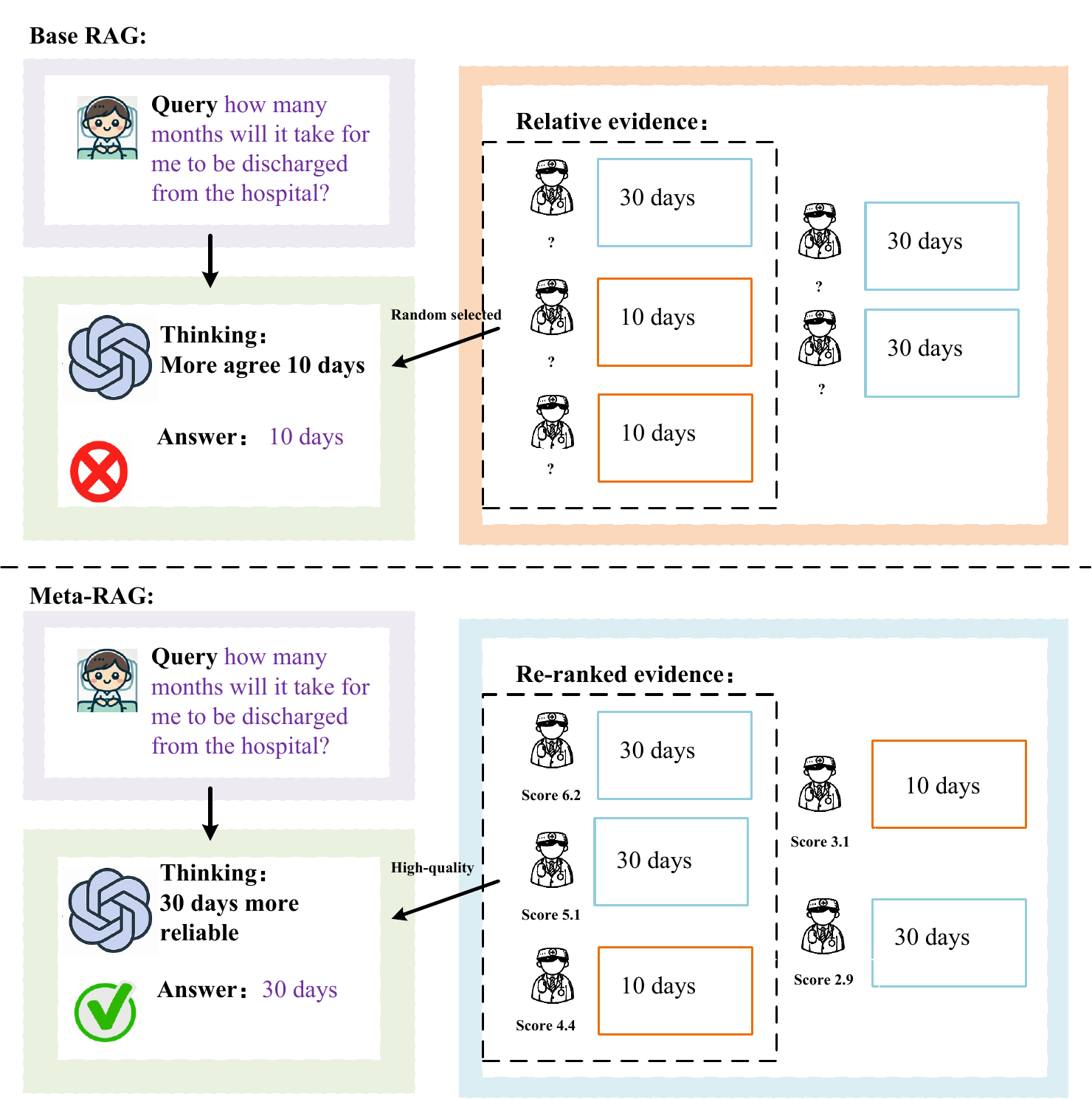}
    \caption{When traditional RAG processes a query, it probably retrieves a large volume of unhelpful and non-professional evidence. This evidence may include conditional results and outdated conclusions. This will mislead the generator to mistakes. }
    \label{fig:1}
\end{figure}
With the iterative advancements in LLM technology, innovative methods like Retrieval-augmented Generation(RAG) and knowledge fine-tuning have emerged~\cite{alam2023automated}. They can minimize the knowledge errors made by LLMs~\cite{zhang2023siren,huang2023survey}. The core process of RAG, which involves retrieving evidence and generating diagnoses, closely aligns with the fundamental principles of EBM. As a result, RAG has the most potential to enhance the efficiency of EBM. However, RAG faces several limitations when applied to clinical medicine. EBM requires a highly rigorous process for selecting and filtering the retrieved evidence~\cite{sackett2008evidence}. Traditional RAG fails to adequately address this process because of the complexity of medical articles. This oversight often leads to the retrieval of conflicting and redundant evidence. For instance, as illustrated in Figure~\ref{fig:1}, the vanilla RAG probably retrieves a large volume of unhelpful and non-reliable evidence. This evidence may include conditional results and outdated conclusions. Consequently, RAG selects this evidence to mislead the response, which will significantly restrict the accuracy.

\begin{figure*}[t]
    \centering
    \includegraphics[width=0.9\linewidth]{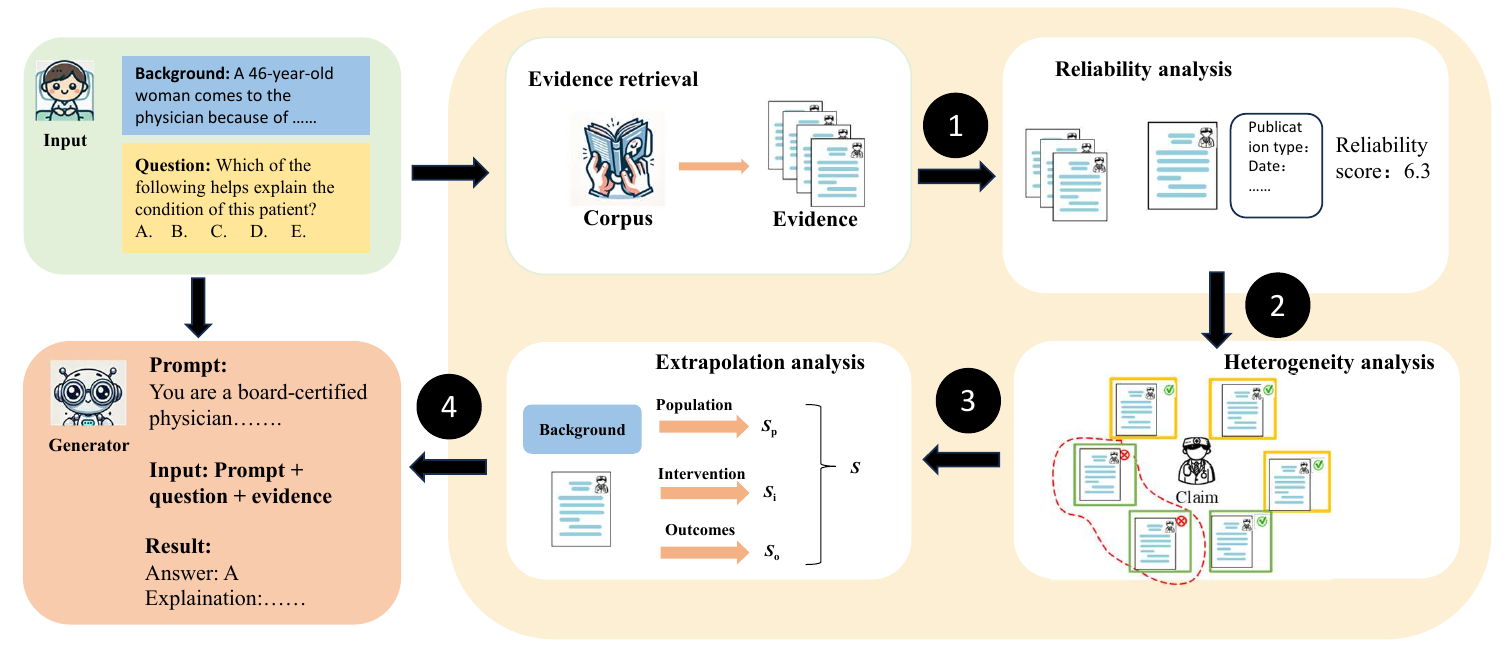}
    \caption{The pipeline of META-RAG includes (1) reliability analysis, (2) heterogeneity analysis, and (3) extrapolation analysis. Our method incorporates these three stages to re-rank and filter evidence, providing as high-quality evidence as possible to (4) generator LLM.}
    \label{fig:3}
\end{figure*}

To address the above issues, we develop META-RAG for evidence re-ranking and filtering in RAG for EBM. By acquiring more reliable and valid evidence, this method enables RAG to retrieve evidence that is both more trustworthy and consistent, thereby reducing erroneous judgments. We emulate the principles of meta-analysis, which focuses on three key aspects: (1) reliability, (2) heterogeneity, and (3) extrapolation~\cite{lipsey2001practical,egger1997meta,hansen2022conduct}. META-RAG filters out inconsistent evidence and presents reliable and rigorous evidence to the response model. As shown in Figure~\ref{fig:3}, first, we gather the related medical articles and assign a base score to each article based on its publication type. Then, we assess the information of evidence to judge the reliability score accordingly and filter the heterogeneous articles. We evaluate the extrapolation by considering the limitations of the experimental results for the users. Finally, the reliable and high-quality articles are passed to the generator. We present the experiments and results to prove our method effectively resolves the issues of low-quality and conflicting evidence. This method can significantly improve the accuracy of the RAG process in providing correct responses.

 Our contributions can be summarized in three aspects: 
\begin{itemize}
    \item  Inspired by meta-analysis, we re-rank the evidence by adopting the evaluation dimensions from meta-analysis, assessing the evidence based on its grade, methodological reliability, and extrapolation.
    \item   We utilize LLM agents to analyze the extrapolative and reliable potential of the evidence, reducing subjectivity in the evidence selection process.
    \item  We conduct an evaluation method for the quality of evidence. By scoring the contribution of the articles to each option, we can observe the improvement of evidence.
\end{itemize}

\section{Related Works}
\subsection{EBM and Meta-Analysis}


Healthcare professionals have recognized and accepted EBM as an important discipline in recent years. EBM aims to make the best clinical decisions by integrating the best research evidence, clinical expertise, and patient preferences~\cite{subbiah2023next,mcmurray2021should}. Many practitioners of EBM attempt to use advanced AI technologies to aid in the search process. However, doctors cannot trust AI models because of hallucinations. They would like to choose the time-consuming and subjective manual approach unless the LLMs~\cite{li2024benchmarking}.


To eliminate biases arising from subjective choices, researchers propose the method known as meta-analysis. Meta-analysis is a quantitative research technique designed to systematically integrate the results of multiple independent studies to provide more rigorous conclusions. It is widely used in fields like medicine, social science, and education, especially in studies derived from experiments~\cite{borenstein2021introduction}. In meta-analysis, researchers aggregate data from multiple independent studies and conduct uniform statistical analyses to determine overall effect sizes or other relevant statistical metrics~\cite{hansen2022conduct}. However, each meta-analysis requires manually compiling more relevant literature, which is highly complex. Therefore, we hope to utilize the core comparative elements of meta-analysis and employ LLMs to assist users in evaluating evidence.

\subsection{RAG for EBM }

LLMs have recently made significant progress in natural language processing. High-performance models like GPT-4~\cite{achiam2023gpt} have achieved substantial breakthroughs in fields such as medicine, military, and law. Google MED-PALM~\cite{singhal2023large} suggests that LLMs can be applied in many tasks within clinical. With RAG method, the LLMs can deal with these complex tasks with few hallucinations~\cite{lewis2020retrieval}. The principle of EBM, which relies on extensive medical evidence for decision-making, aligns well with this approach. RAG generative method is particularly well-suited for EBM and serves as an effective tool for assisting doctors in resolving clinical issues.

However, medicine constantly evolves at a rapid pace, leading to inconsistencies in viewpoints among publications like the articles in PubMed~\cite{white2020pubmed}. RAG may retrieve outdated, incorrect, and restricted theories. They may have once been accepted but no longer right because of the proposal of a new theory. This phenomenon will result in some conclusions being inapplicable to the actual situation. 

\subsection{Evidence Re-Ranking }


Currently, there are three main methods for optimizing the evidence retrieved during the RAG process: scoring based on rules, trained models, and LLMs that have re-ranking capabilities~\cite{gao2023retrieval}. Researchers tend to employ existing rules for the task, relying on predefined metrics such as diversity, relevance, and Mean Reciprocal Rank (MRR)~\cite{gao2023retrieval}. By calculating specific values for these articles, those with higher values are prioritized in the ranking. Model-based approaches used traditional Transformer models like SpanBERT~\cite{joshi2020spanbert}. 

The third method utilizes some specialized re-ranking models like Cohere re-rank or be-ranked-large and general-purpose LLMs like GPT~\cite{gao2023chat}. Filtering evidence also effectively optimizes evidence quality. There is another Filter-re-ranker paradigm combining the strengths of different models~\cite{ma2023large}. The smaller model acts as a filter while the LLM serves as a re-ranking agent. Another simple and effective method involves LLMs evaluating the retrieved content before generating the final answer, allowing the LLM to self-assess and filter out documents with poor relevance. 


\section{Method}
\subsection{Task Definition}
To align with the principles of EBM, we aim not only to deliver convincing answers but also to present high-quality evidence. We define medical queries $Q$ from users as system inputs and then respond $A$ and retrieved evidence $E$ as the output. As shown in Figure \ref{fig:3}, our main pipeline focuses on the re-ranking and filtering steps of the evidence in RAG. At the end of the re-ranking and filtering section, we pass high-quality articles with their orders to the generator. In this task, we evaluate the evidence across three distinct dimensions: reliability analysis, heterogeneity analysis, and extrapolation analysis. These analyses enable us to assess the reliability of the evidence, exclude untrustworthy findings, and determine whether the results can be applied to the patient. After re-ranking evidence, the most effective pieces of evidence and their order are passed to the response model to generate recommendations for the queries.

\subsection{Evidence Retrieval}


In the first step, we conduct evidence retrieval based on query similarity with the datasets.
However, there are too many article types in PubMed~\cite{white2020pubmed}. A substantial proportion of the articles lack an abstract. To address these problems, we employ a hybrid retrieval approach. We simultaneously search the article titles, abstracts, and MeSH (Medical Subject Headings) keys in the articles. By calculating and aggregating the similarity scores across these three different tags and then ranking them, we ultimately select the evidence $E$ with the highest scores as potential evidence.

\subsection{Reliability Analysis }

\begin{figure}[h]
    \centering
    \includegraphics[width=0.8\linewidth]{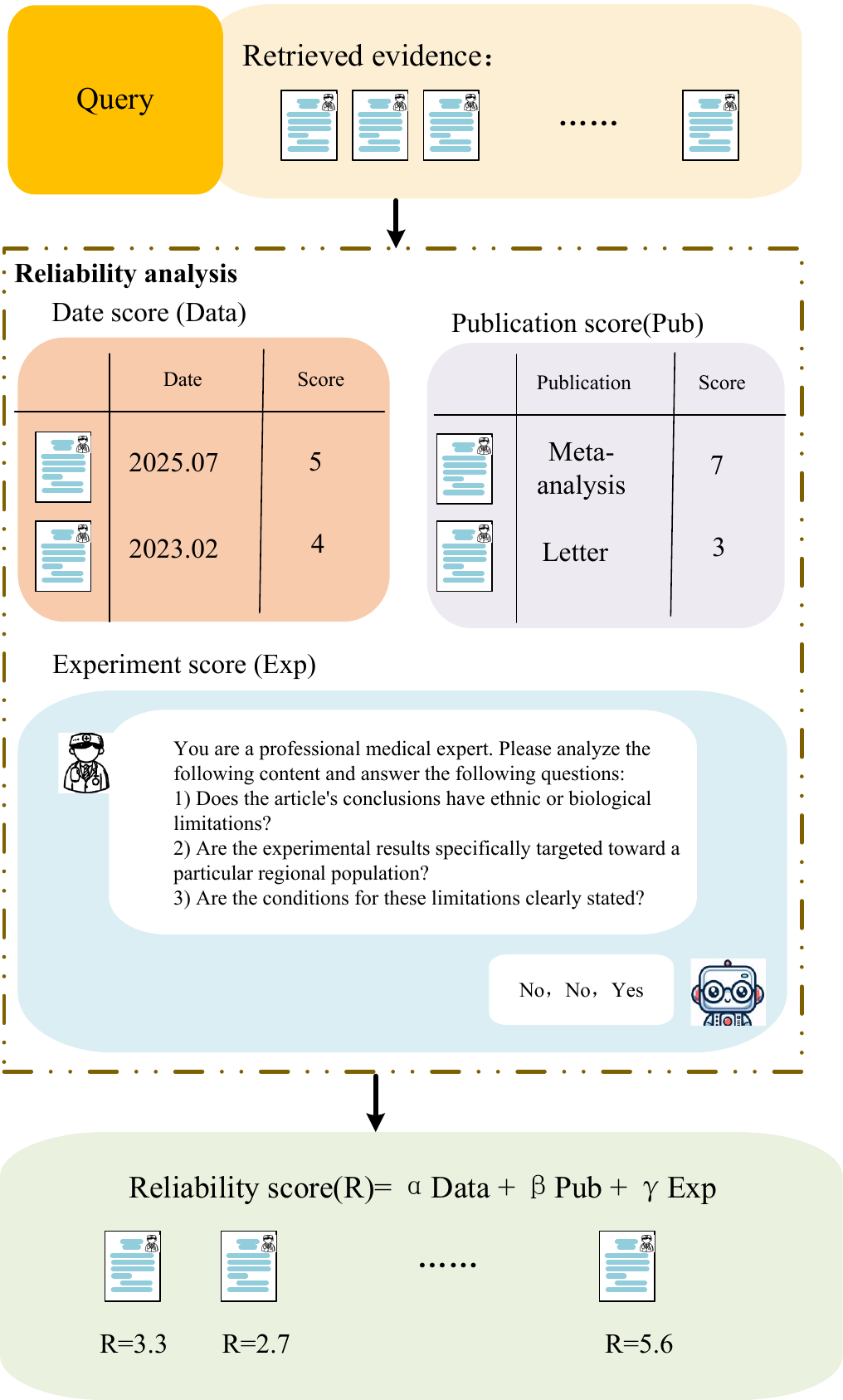}
    \caption{The pipeline of the reliability analysis. We synthesize the information and the judgments of LLM to show the reliability of each evidence.}
    \label{fig:5}
\end{figure}

After obtaining highly relevant evidence, we first grade the articles by their fundamental information. As shown in Figure~\ref{fig:5}, we mainly score the evidence $E$ with the rules of the publication type, publication date, and LLM judgments. Initially, we access the publication type from the information and evaluate the evidence quality level. We assign scores ranging from 1 to 7 based on the medical principles~\cite{polit2004nursing}. Recognizing that the publication date of an article can significantly influence its conclusions, we then sort the articles by their publication dates. We award an extra point to the most recently published articles on their base score. And as the article becomes less recent, the score we reward gradually decreases in tiers.
 This process results in our base score derived from rule-based filtering.

We also employ an LLM for a more fine-grained reliability analysis. Meta-analyses typically analyze the randomization of literature, data integrity, presence of bias, and choices regarding blinding. These principles can reflect the validity of the experimental conclusions in the article. We implement this method, evaluating the evidence by three questions as detailed in Figure~\ref{fig:5}. The detailed 
architecture of prompts and questions is provided in the appendix. Ultimately, this process provides the reliability score $r_i$ with $E_i$. A larger $r_i$ signify a more rigorous methodology.

\subsection{Heterogeneity Analysis }

After we score each piece of evidence on reliability, we then apply heterogeneity analysis. This analysis can remove studies with low quality and high heterogeneity. This step guarantees that only valid evidence is fed to the generative model. We perform heterogeneity detection for each article-claim pair. In this analysis, we apply the definition of heterogeneity in the DerSimonian-Laird method~\cite{dersimonian2015meta} to filter the evidence. Based on the characteristics of this dataset, we approximate part of the model parameters and define the measurement metric to represent the stance of each article. \textbf{First}, we create claims by combining the query with each option. Each option defines a separate claim. We ask LLMs to determine the stance of each piece of evidence on each claim. We define the label of each evidence as $y_i$, and mark these pairs as support, oppose, or irrelevant.

\begin{equation}
y_i \;=\;
\begin{cases}
1, & \text{if } i~\text{labeled “Support”},\\
0, & \text{if } i~\text{labeled “Oppose”},\\
\text{NaN}, & \text{if}~i~\text{labeled “Irrelevant”}.
\end{cases}
\end{equation}
Then we need to compute the heterogeneity of the evidence set associated with each query. We compute the random-effects variance $\tau^2_{DL}$, the pooled effect $\theta_{RE}$, and the study weights $w_{re}$ at this step. We define $k$ as the total number of studies retrieved for a single query and $v_i$ as the variance estimate of the $ i^ {th} $ study. Most original studies do not report standard errors, so we set $v_i$ as $\mathrm{\sigma}$. Define $w_i$ as the weight of the $ i^ {th} $ observation in the fixed‐effect model. Then compute the fixed‐effect combined estimate $\hat\theta_{\mathrm{FE}}$.  Formally,

\begin{equation}
\hat\theta_{\mathrm{FE}}
\;=\;
\frac{\sum_{i=1}^k w_i\,y_i}
     {\sum_{i=1}^k w_i}
\quad\text{where}\quad
w_i \;=\;\frac{1}{v_i}
\end{equation}

Afterwards, we calculate the heterogeneity statistic $Q$. This variable represents the total standard deviation of the entire set of articles. It serves as a preliminary indicator of the consistency of stances within the article cluster.
 
\begin{equation}
Q
\;=\;
\sum_{i=1}^k
w_i\,\bigl(y_i - \hat\theta_{\mathrm{FE}}\bigr)^2
\end{equation}
Then calculate the heterogeneity $\tau^2_{\mathrm{DL}}$ under the DerSimonian–Laird random‐effects model.
These metrics measure the fixed-effect estimate dispersion and the true effect variability across studies. Formally,

\begin{equation}
\tau^2_{\mathrm{DL}}
\;=\;
\max\!\Biggl\{
\frac{Q - (k - 1)}
     { \sum_{i=1}^k w_i \;-\;\frac{\sum_{i=1}^k w_i^2}{\sum_{i=1}^k w_i}}
\;,\;0
\Biggr\}
\end{equation}

We use $\tau^2_{\mathrm{DL}}$ and $v_i$ to derive the random-effects weight $W_i$ and the overall estimate $\hat\theta_{\mathrm{RE}}$. We also compute each study’s outlier measure $Q_i$ and the leave-one-out heterogeneity reduction ratio $\Delta_i$. Formally,

\begin{equation}
\hat\theta_{\mathrm{RE}}
\;=\;
\frac{ \sum_{i=1}^k W_i\,y_i}
     { \sum_{i=1}^k W_i}
\quad\text{where}\quad
W_i \;=\;\frac{1}{v_i + \tau^2_{\mathrm{DL}}}
\end{equation}

\begin{equation}
Q_i
\;=\;
\frac{\bigl(y_i - \hat\theta_{\mathrm{RE}}\bigr)^2}{v_i}
\end{equation}

Finally, we define $S$ as the set of all $k$ studies and $S^{(-i)}$ as the set obtained by removing study $i$ from $S$. As for the formula 4, we compute the $\tau_{\mathrm{DL}}^{2\,(-i)}$ for the $S^{(-i)}$ and calculate the decrease caused by this evidence. We define an acceptable maximum heterogeneity contribution $M$ and a minimum reliability score $R_c$. Based on the final outcomes, we determine whether each article should be excluded. Formally,

\begin{equation}
\label{eq7}
\Delta_i
\;=\;
\frac{\tau^2_{\mathrm{DL}}\;-\;\tau^{2\,(-i)}_{\mathrm{DL}}}
     {\tau^2_{\mathrm{DL}}}
\end{equation}

Algorithm ~\ref{algr-Construction} summarizes this process.

\begin{algorithm}[!h]
\caption{Heterogeneity Analysis }
\label{algr-Construction}
\textbf{Input:} Query $q$, Evidence $(E,R) = \{(E_1,r_1), \dots \, (E_k,r_k)\}$, Hyperparameter $M,R_c$ \\
\textbf{Output:} filtered evidence $E_f = \{E_1, \dots \, E_m\}$
    \begin{algorithmic}[1]
    \State  $v_i \gets \sigma $  \Comment{Initialize}
    \State ${c_1,c_2,\dots} \gets \mathcal{C}(q)$ \Comment{Combine claims}
    \State $ E_f \gets \{\}$
    \For{$ c_i \in \mathcal{C}(q)$}
        \State $y \gets \mathcal{G}(\mathcal{P}_0(c,E_i))$ \Comment{Generate evidence labels}        \State $\tau^2_{\mathrm{DL}} \gets \max{D}(q, y,v_i, k)$ \Comment{Calculate DL variance}
        \For{$e \in \{E_i \mid i=0 \dots k \}$}
            \State Compute $Q_i = \mathcal{M}(y_i,v_i)$ 
            \State Compute $\Delta_i$ by Eq.~\eqref{eq7}
            
            \If {$\Delta_i < M \;\land\; e\notin E_f$}
                \State Add $e$ to $E_f$ 
            \EndIf
            \If {$\Delta_i \ge M \;\land\; e\notin E_f \;\land\; r_i>R_c$ }
                \State Add $e$ to $E_f$ 
            \EndIf
        \EndFor
    \EndFor
    \State \textbf{return} $E_f$ \Comment{Return the filtered evidence}
    \end{algorithmic}
\end{algorithm}

\subsection{Extrapolation Analysis }

To prevent large gaps between the background of the user and the experimental conditions in the evidence, we adjust the extrapolation score of each evidence. This process is implemented based on the similarity between them. We divide the process into three clear steps. First, we split the query into the user background and the clinical question. Then, we use LLM with a carefully designed prompt to compare the background information from the query and the evidence across the population, intervention, and outcomes(PIO)~\cite{methley2014pico}. Through this process, each piece of evidence is assigned a fine-grained score along each of these dimensions, and the detailed architecture of this process is provided in the appendix. Finally, we compute an overall extrapolation score for each evidence relative to the user’s background. We calculate the final ranking score $S$ by both the extrapolation score and the reliability score.

Algorithm ~\ref{algr-Extra} summarizes this process.

\begin{algorithm}[!h]
\caption{Extrapolation Analysis}
\label{algr-Extra}
\textbf{Input:} Query $q$, Evidence $E_f,R_f = \{(E_1,r_1), \dots \, (E_f,r_f)\}$, Hyperparameter $\alpha,\beta,\gamma$ \\
\textbf{Output:} Scored evidence $(E_f,S) = \{(E_1,S_1), \dots \, (E_m,S_m)\}$
    \begin{algorithmic}[1]
    \State $S\gets \{\}$ \Comment{Initialize}
    \State ${Back,Que} \gets \mathcal{C}(q)$ \Comment{split the background}

    \For{$e \in \{E_j \mid j=0 \dots m\}$}
        \State $T_p \gets \mathcal{G}(\mathcal{P}_0(Back,E_j))$ \Comment{Generate Population score}
        \State $T_i \gets \mathcal{G}(\mathcal{P}_1(Back,E_j))$ \Comment{Generate Intervention score}
        \State $T_o \gets \mathcal{G}(\mathcal{P}_2(Back,E_j))$ \Comment{Generate outcome score}
        \State $T_j\gets \alpha T_p+\beta T_i + \gamma T_o$ \Comment{Calculate Extrapolation score}
        \State $S_j\gets r^2_jT_j$ \Comment{Calculate total ranking score}
    \EndFor
    \State \textbf{return} $(E_f,S)$ \Comment{Return the filtered evidence}
    \end{algorithmic}
\end{algorithm}

\section{Experiments and Results}
\subsection{Experiments Setup}

\textbf{Datasets} In our experiment, we first select various medical Q\&A datasets and literature databases for our query resource. To ensure our experimental results are clear and fair, we select a five-option multiple-choice Q\&A dataset as the task format. During data selection, we guarantee that each question includes sufficient patient information to support extrapolation analysis. We ultimately focus our method on the MedQA~\cite{jin2020disease} and MMLU~\cite{he2019applying}, which contain more comprehensive and professional user queries. The MedQA dataset typically comprises real cases that carry patient information, allowing us to perform extrapolative analysis more accurately. 

\begin{figure}[t]
    \centering
    \includegraphics[width=0.9\linewidth]{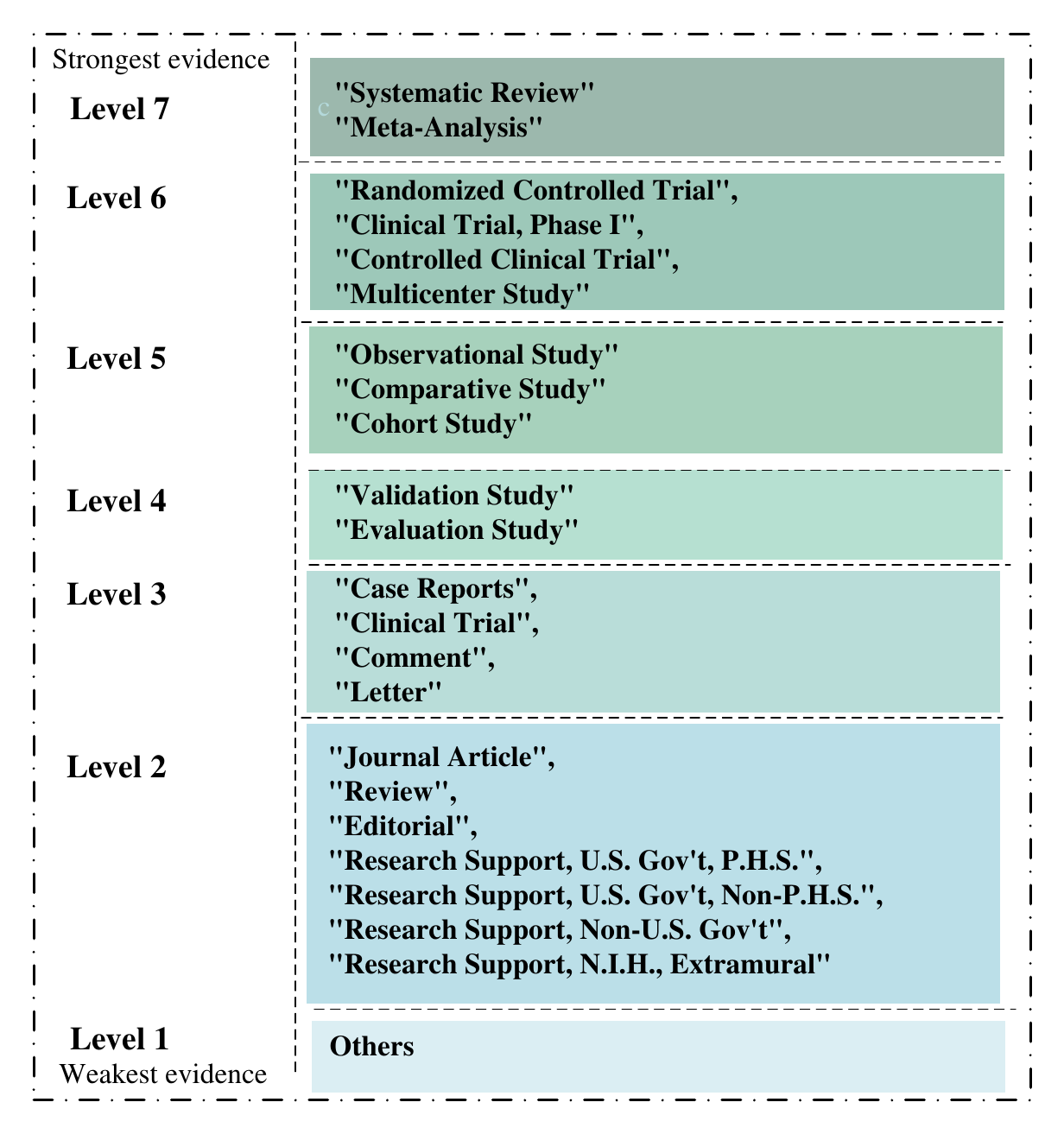}
    \caption{We divide the evidence type into 7 levels. In reliability analysis, we categorize evidence from different publication types and LLM judgments. The higher level of evidence means a better publication type score.}
    \label{fig:4}
\end{figure}

\begin{figure}[t]
    \centering
    \includegraphics[width=0.9\linewidth]{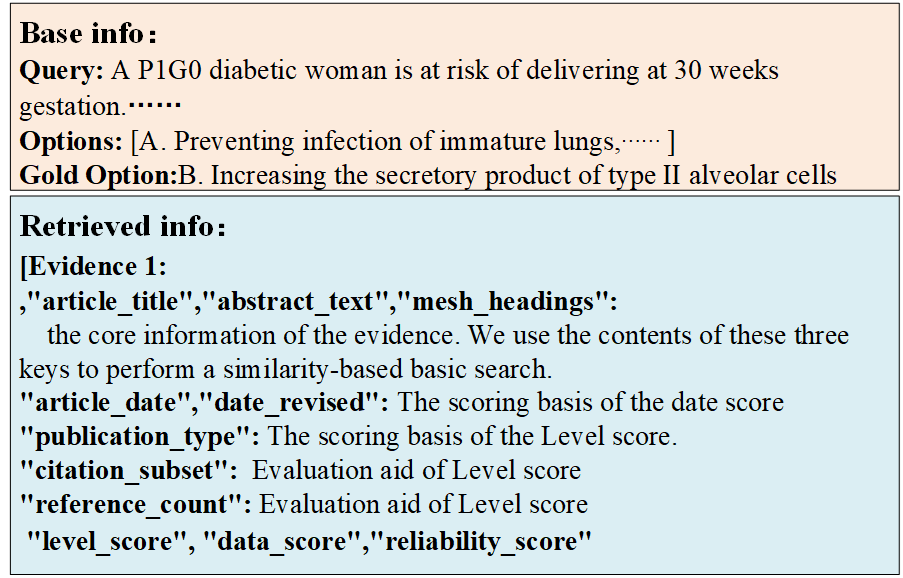}
    \caption{The specific information structures retrieved from the PubMed dataset. By analyzing this detailed information of articles, we can comprehensively assess whether the literature is sufficiently authoritative.}
    \label{fig:8}
\end{figure}

\textbf{Evidence corpus} Also, we take the PubMed dataset as the literature database, which is widely used in medical meta-analysis work. This dataset provides a thorough organization of information from the literature. As shown in Figure~\ref{fig:8}, the PubMed data set provides all the information we need for better retrieval and evaluation. In the step of reliability analysis, we also divide these articles into different levels by the rules shown in Figure~\ref{fig:4}. The top articles have the strongest evidence grade due to their publication type. Because the classification of PubMed articles is more detailed compared to traditional medical evidence grades, the order in our ranking is based on a fusion of multiple medical field evidence grading systems. Our ranking primarily follows the ~\cite{polit2004nursing} method, categorizing these articles into seven levels.

\subsection{Main Result}
\begin{table*}[h!]
    \centering
    \small
    \renewcommand{\arraystretch}{1.0} 
    \begin{tabular}{cp{1.5cm}|p{0.6cm}p{0.6cm}p{0.6cm}p{0.6cm}p{0.6cm}|p{0.6cm}p{0.6cm}p{0.6cm}p{0.6cm}p{0.6cm}}
        \toprule
        \toprule
        \multicolumn{2}{c|}{\multirow{2}{*}{\textbf{method}}} & \multicolumn{5}{c|}{\textbf{MedQA}} & \multicolumn{5}{c}{\textbf{MMLU}} \\
        & & D1 & D2 & D3 & D4 & Best & D1 & D2 & D3 & D4 & Best \\
        \midrule
        \multirow{4}{*}{Llama-3.0-8B} & Meta & \textbf{44.0} & 38.0 & 40.7 & 39.3 & \textbf{44.0} & \textbf{42.7} & 42.0 & 42.0 & 39.3 & \textbf{42.7}\\
         & w/o Evi & - & - & - & -& 38.7 & - & - & - & - &36.2 \\
         & Ran-Evi & 25.3 & 29.3 & 31.9 & 32.6 & 32.6 & 25.3 & 29.3 & 31.9 & 32.6& 32.6 \\
         & Self-Evi & 38.0 & 30.0 &33.3 & 28.3 & 38.0& 36.0 & 40.0 & 42.7 &37.2& 42.7 \\
         \midrule
        \multirow{4}{*}{Qwen2.5-7B} & Meta & 51.5 & \textbf{52.0} & 48.5 & 42.5 & \textbf{52.0} & 49.3 & 46.0 & \textbf{50.7}  & 46.7& \textbf{50.7}\\
         & w/o Evi & - & - & - & - &49.6 & - & - & - & -& 49.3 \\
        & Ran-Evi & 44.5 & 43.5 & 42.5& 43.5 & 44.5 & 43.3 & 43.7 & 48.4& 44.3& 48.4 \\
         & Self-Evi &42.5 & 39.5 & 43.5 & 41.5 & 41.5 & 48.0 & 48.7 & 48.0  & 48.4& 48.7\\
         \midrule
         \multirow{4}{*}{Mistral-7B} & Meta & \textbf{47.5 }& 45.0 & 46.5 & 46.5 & \textbf{47.5} & 45.0 & 47.3 & \textbf{48.0}  & 47.7& \textbf{48.0}\\
         & w/o Evi & - & -& - & - &43.5 & - & - & -  & -& 44.0\\
         & Ran-Evi &42.0 & 42.5 & 40.5& 45.5 & 45.5 & 43.3 & 45.0 & 46.8  & 45.4& 46.8\\
         & Self-Evi & 42.5 & 39.5 & 43.5 & 41.5 & 43.5 &43.3 & 44.7 & 45.3 & 46.7& 46.7\\
         \midrule
         \multirow{4}{*}{Gemma-1.1-7B} & Meta & 41.0 & 41.5 & \textbf{43.0} & 40.0 & \textbf{43.0} & 36.0 & 34.7 & 35.3 & \textbf{40.0}& \textbf{40.0} \\
         & w/o Evi & - & - & - & - & 40.5 & - & - & -  & -& 34.7\\
        & Ran-Evi & 34.0 & 31.5 & 30.0 & 31.0 & 34.0& 35.3 &36.6 & 35.5  & 37.1& 37.1\\
         & Self-Evi & 31.0 & 29.5 & 30.0 & 31.0 & 31.0 & 34.7& 29.3 & 33.3 & 34.7& 34.7 \\

        \bottomrule
         \bottomrule
    \end{tabular}
    \caption{Accuracy (\%) of Meta-RAG and other baselines in 300 queries of MedQA and MMLU datasets. w/o Evi: unrelated evidence provided; Ran-Evi: evidence randomly selected by correlation; Self-Evi: evidence selected by the generator LLM. All the other LLMs below use the re-rank method like Self-Evi. The numbers D1, D2, D3, and D4 under the dataset name represent the number of evidence articles provided to the model during generation.}
    \label{tab:1}
\end{table*}

\begin{table}[h!]
    \centering
    \small
    \renewcommand{\arraystretch}{1.0} 
    \begin{tabular}{cp{0.9cm}|p{0.5cm}p{0.5cm}p{0.5cm}p{0.5cm}p{0.6cm}}
        \toprule
        \toprule
        \multicolumn{2}{c|}{\multirow{2}{*}{\textbf{method}}} & \multicolumn{5}{c}{\textbf{MedQA}}  \\
        & & D1 & D2 & D3 & D4 & Best\\
        \midrule
        \multirow{4}{*}{0.5B} & Meta & \textbf{25.00} & 24.70 & 23.92 & 23.88 & \textbf{25.00} \\
         & w/o & - & - & - & -& 24.64  \\
         & Ran- & 23.80 & 23.50 &23.62 & 23.80 & 23.80 \\
         & Self- & 23.50 & 23.78 & 24.00 & 23.34 & 24.00  \\
         \midrule
        \multirow{4}{*}{1.5B} & Meta & 28.42 & 30.26 & 30.56 & \textbf{30.82} & \textbf{30.82} \\
         & w/o & - & - & - & - & \textbf{35.08} \\
        & Ran- &26.52 & 28.84 & 28.12 & 27.40 & 28.84  \\
         & Self- &25.98 & 28.48 & 27.94 & 27.08 & 28.48\\
                  \midrule
        \multirow{4}{*}{7B} & Meta & 50.76 & \textbf{51.04} & 50.74 & 50.56 & \textbf{51.04} \\
         & w/o & - & - & - & - &50.58 \\
        & Ran- &47.04 & 47.08 & 46.66 & 47.18 & 47.18  \\
         & Self- &49.70 & 49.12 & 49.52 & 49.18 & 49.70 \\
                 \midrule
        \multirow{4}{*}{14B} & Meta & 59.20 & 60.00 & 60.72 & \textbf{60.90} & \textbf{60.90} \\
         & w/o & - & - & - & - &58.36 \\
        & Ran- &56.58 & 56.76 & 56.32 & 56.94 & 56.94  \\
         & Self- &55.48 & 55.90 & 55.88& 55.66 & 55.90\\
                  \midrule
        \multirow{4}{*}{32B} & Meta & 63.28 & 64.08 & 64.00 & \textbf{64.32} & \textbf{64.32} \\
         & w/o & - & - & - & - &62.06 \\
        & Ran- & 59.14 & 60.24 & 60.30 & 60.10 & 60.30 \\
         & Self- &59.46 & 60.24 & 60.10 & 60.18 & 60.24\\

        \bottomrule
         \bottomrule
    \end{tabular}
    \caption{Accuracy (\%) of the responses of different sizes of Qwen-2.5 LLMs. All the responses are based on 5000 queries of the MedQA datasets. }
    \label{tab:2}
\end{table}
 We select three different baselines to show the performance of the META. The experiments are calculated based on 5000 queries extracted from the MedQA datasets and 300 queries extracted from MMLU and MedQA datasets. Due to limited computational resources, all the evidence used in the experiments is extracted from over four million medical articles in PubMed. For each query, 15 articles are initially retrieved as a baseline, and then different methods are used for filtering and re-ranking.

\textbf{w/o Evi}. To test the base performance of each model, we give no evidence to the LLM as w/o Evi. This baseline can test whether the LLM has studied this query. Actually, some LLMs like GPT-4o-mini reach an accuracy of over 90\%. This performance may have no increase by the META method.

\textbf{Ran-Evi}. We provide evidence extracted based on the LLM as Ran-Evi. We use a random function to shuffle the extracted evidence, serving as the most straightforward control group for our method.

\textbf{Self-Evi}. We provide a random order of evidence as the Self-Evi. This baseline is designed to demonstrate that our method offers a significant improvement over traditional LLM-based re-ranking approaches. We utilized the inherent capability of each large model to rank the relevance of documents in the evidence pool. The top-ranked documents from this sorted list were then selected as evidence and provided to the generation model. 

Experimental results are presented in Table~\ref{tab:1} and Table~\ref{tab:2}. Table~\ref{tab:1} shows the performance of different types of LLMs, while Table~\ref{tab:2} shows the performance of different sizes. We can get the following analysis:

1) Our method consistently improves upon the baselines of almost all LLMs. Across LLMs of different sizes and types, our Meta-RAG achieves substantial improvements over traditional evidence-ranking methods.

2) In the response by Qwen-2.5, we observe a clear accuracy gain when the model answers directly. In the $1.5\mathrm{B}$ model, w/o Evi reaches an accuracy of 35.08\%. Both Ran-Evi and Self-Evi, which include additional evidence, perform worse than w/o Evi. One possible explanation for this phenomenon is that the Qwen model has memorized this particular question, so the extra input tokens harm the performance. For other differently sized models, however, our high-quality evidence still yields improved accuracy.

3) When given different amounts of evidence, our method delivers steadily increasing performance on stronger models. Small models suffer significant drops when exposed to too many input tokens. As a result, Table 1 shows large overall performance fluctuations. However, for Qwen-14B and Qwen-32B, adding more evidence consistently improves accuracy, indicating that our evidence has low heterogeneity.

\subsection{Ablation Study}
To validate the efficiency of each step, we set multiple ablation experiments. For the reliability ablation experiments, we ablate the reliability analysis module of the model. We set all the reliability scores sent to the heterogeneity analysis same. When calculating the highest-scoring evidence for heterogeneity analysis, we randomly select the first piece of evidence from the list. After conducting this ablation, we observe that both the quality of the evidence and the accuracy of the responses decrease, as shown in Table~\ref {tab:3}.

\begin{table}[htbp]
    \begin{center}
    \renewcommand{\arraystretch}{1.0} 
    \begin{tabular}{p{1.0cm}|p{0.6cm}p{0.6cm}p{0.6cm}p{0.6cm}p{0.6cm}}
        \toprule
        \toprule
        \multicolumn{1}{c|}{\multirow{2}{*}{\textbf{method}}} & \multicolumn{5}{c}{\textbf{MedQA}}  \\
        & 1 & 2 & 3 & 4 & Best  \\
        \midrule
          Meta & \textbf{44.0} & 38.0 & 40.7 & 39.3 & \textbf{44.0}\\
          \textit{w/o R} & 36.7 & 40.0 & 34.0 & 37.3& 40.0 \\
          \textit{w/o H} & 34.0 & 38.7 & 37.3 & 35.3 & 38.7  \\
          \textit{w/o E} & 34.0 & 33.3 &34.7 & 34.7& 34.7\\
        \bottomrule
        \bottomrule
        \end{tabular}
        \caption{Ablation of each part of the Meta \textit{\textbf{w/o R}} means no reliability checking and all the reliability scores set to 1. \textit{\textbf{w/o H}} means no heterogeneity analysis.\textit{\textbf{w/o E}} means no extrapolation analysis}
    \label{tab:3}
    \end{center}
\end{table}
Subsequently, we analyze the impact of heterogeneity on the model. As shown in Table~\ref{tab:3}, we remove the heterogeneity judgment process and directly re-rank the extracted evidence based on reliability and extrapolation. We observe a noticeable decline in the evidence contribution score and accuracy. This phenomenon suggests that some highly reliable but paradoxical evidence scores remain in the evidence and mislead the judgment of the generation model. The generation model then produces responses closer to the incorrect options. Since our calculation of evidence grades involves selecting both the evidence grade and the similarity of the evidence to each option for assessment, this aspect inevitably becomes affected when heterogeneity scoring is removed.

\subsection{Is the Evidence Better?}
\begin{figure}[t]
    \centering
    \includegraphics[width=1.0\linewidth]{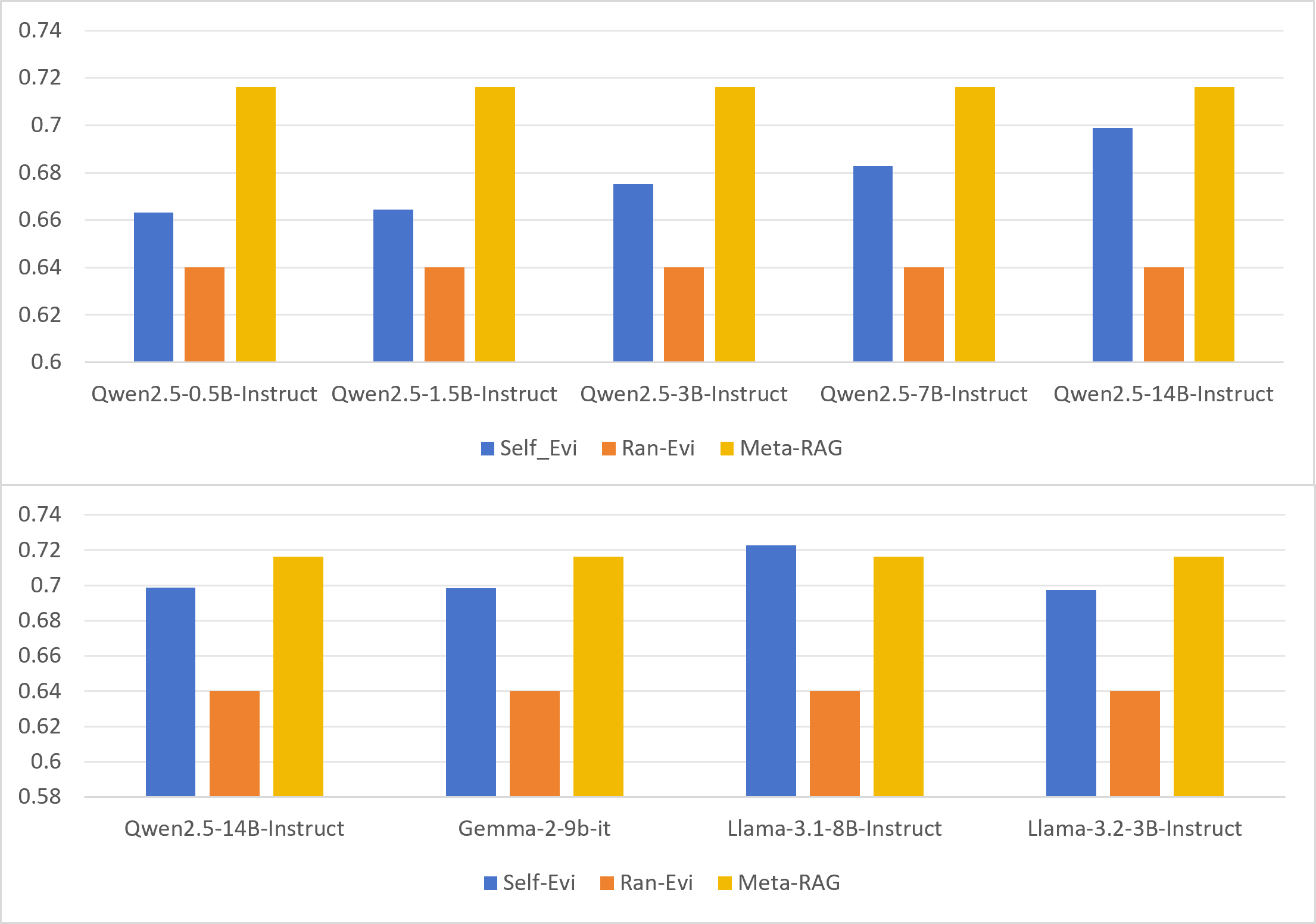}
    \caption{The similarity of each method between the provided evidence and the ground-truth answer. We use this metric to evaluate whether Meta-RAG can better guide the model to the correct answer.}
    \label{fig:9}
\end{figure}

As shown in Figure~\ref{fig:9}, we employ another evaluation metric to evaluate the evidence quality. We assess the similarity of our input evidence and the claim with the gold option to show the contribution to a correct answer. The higher similarity means better evidence is provided. We observe that the average quality of the evidence is effectively enhanced after Meta-RAG. Additionally, we can also analyze that as the model size grows, the Self-Evi group becomes more sensitive to good evidence. In addition, the self-ranking ability varies between LLMs. Our experiments show that some models can select higher-quality evidence. Yet they do not achieve higher accuracy when using self-selected evidence. We believe that this happens because, although the chosen evidence is more similar to the correct answer, it is also controversial. That controversy raises the similarity to incorrect answers. As a result, the generative model becomes confused.
Therefore, most baselines cannot surpass Meta-RAG.

\section{Conclusion}
EBM currently needs robust automated tools to assist in medical tasks. However, existing RAG for EBM cannot ensure the evidence meets the stringent requirements of medicine. Therefore, inspired by the principles of meta-analysis, we propose a META-RAG filtering and re-ranking method to ensure the evidence is effective and reliable. We conduct practical experiments on our method and verify its improvements in accuracy and evidence quality. We hope this work will assist researchers in the medical field, promoting safer and more effective deployment of LLMs in medical applications.

\newpage


\bibliography{aaai2026}

\end{document}


\section{Appendix}
\subsection{Retrieval}
In the first step, we conduct evidence retrieval based on query similarity with the datasets. We aim to retrieve as much relevant evidence as possible. Therefore, we utilize the PubMed database, a vast repository of biomedical and life sciences research articles managed by the U.S. National Library of Medicine. This database primarily consists of academic articles from plenty of publications with extensive information, like mesh heading and the article date.

During the retrieval, we observe that many PubMed entries lack an abstract due to their publication format, so relying solely on abstract‐based retrieval is incomplete. Moreover, selecting articles using only title similarity fails to capture other facets of the claim. To address these limitations, we employed three distinct retrieval strategies and then aggregated and deduplicated their results. For each candidate article, we extracted its abstract, title, and keywords, and computed similarity scores between each of these elements and the claim. We then selected the top ten articles per dimension—thirty in total—and deduplicated this set. Finally, we randomly sampled ten articles from the deduplicated pool and added them to the query’s document directory as our ten extracted pieces of evidence.
\subsection{Reliability Analysis }


After obtaining highly relevant evidence, we first grade the articles by their fundamental information. We combine rules and LLMs judgments to score their reliability. Initially, we access the Publication type from the information and evaluate the articles' quality level. We assign scores ranging from 1 to 7 based on the medical principles~\cite{polit2004nursing}. Recognizing that the publication date of an article can significantly influence its conclusions, we then sort the articles by their publication dates. We award an extra point to the most recently published articles on their base score. And as the article becomes less recent, the score we reward gradually decreases in tiers.
 This process results in our base score derived from rule-based filtering.

Subsequently, we employ an LLM for a more fine-grained reliability analysis. Meta-analyses typically analyze the randomization of literature, data integrity, presence of bias, and choices regarding blinding. These four principles can reflect the validity of the experimental conclusions in the article. We implement this method with an LLM, evaluating the evidence from these four aspects as detailed in Figure 4. We give the LLM a prompt to let it judge the article and respond only four letters to answer the questions.  For each imperfection identified, we apply a penalty parameter to the article score. Ultimately, this process provides us with a more reliable ranking order.
\begin{figure}[h]
    \centering
    \includegraphics[width=1.0\linewidth]{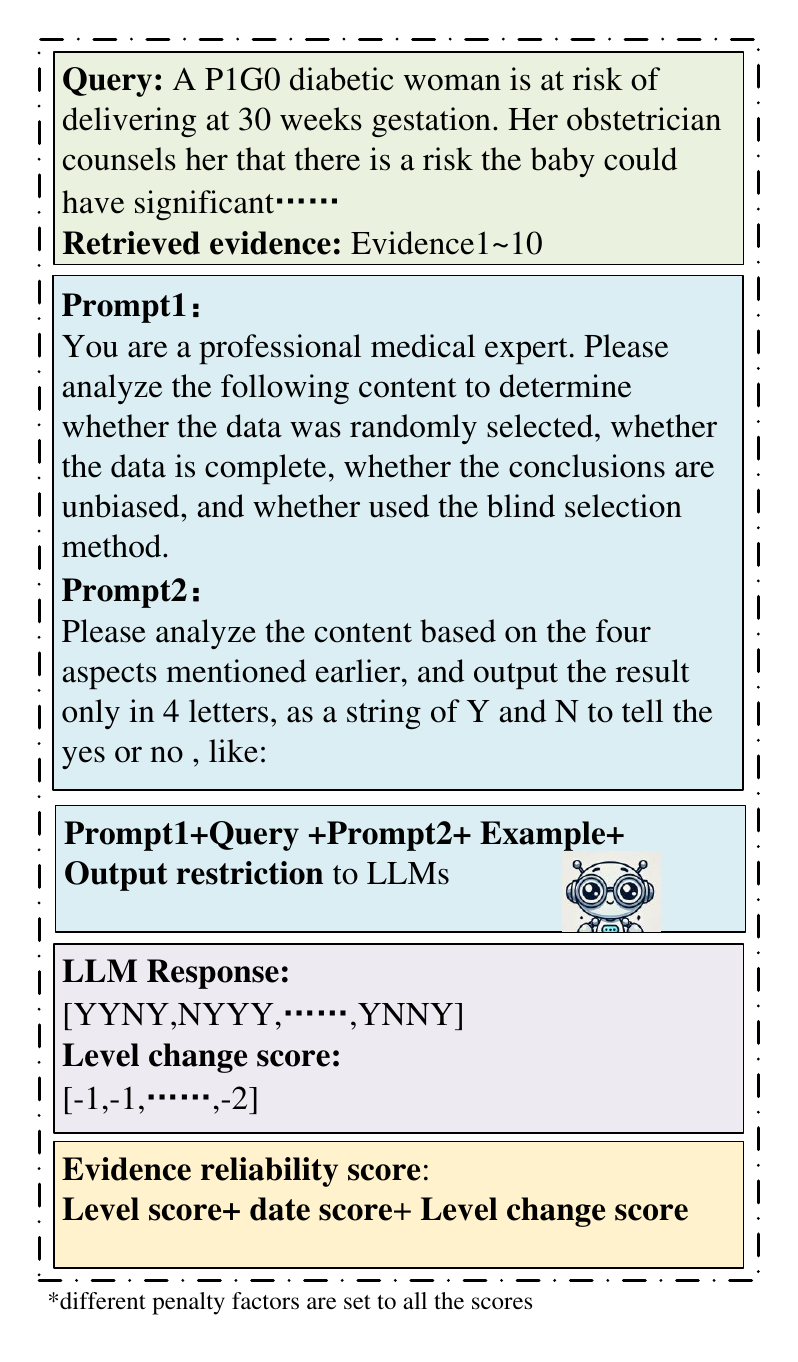}
    \caption{Details of the reliability score counting. This step ensures the articles are ranked by their quality.  }
    \label{fig:5}
\end{figure}
\subsection{Heterogeneity Analysis }

After obtaining the reliability scores, we need a heterogeneity analysis to ensure the content provided to the generation model is closely aligned with the scientific consensus. The traditional meta-analysis need all experiment data to calculate the heterogeneity among all articles. However, we can only access the abstracts of each medical publication, and the raw data of the medical articles is always Closed-source data. Therefore, our approach is to filter out heterogeneous information only based on statistical reliability scores.
\begin{figure}[h]
    \centering
    \includegraphics[width=1.0\linewidth]{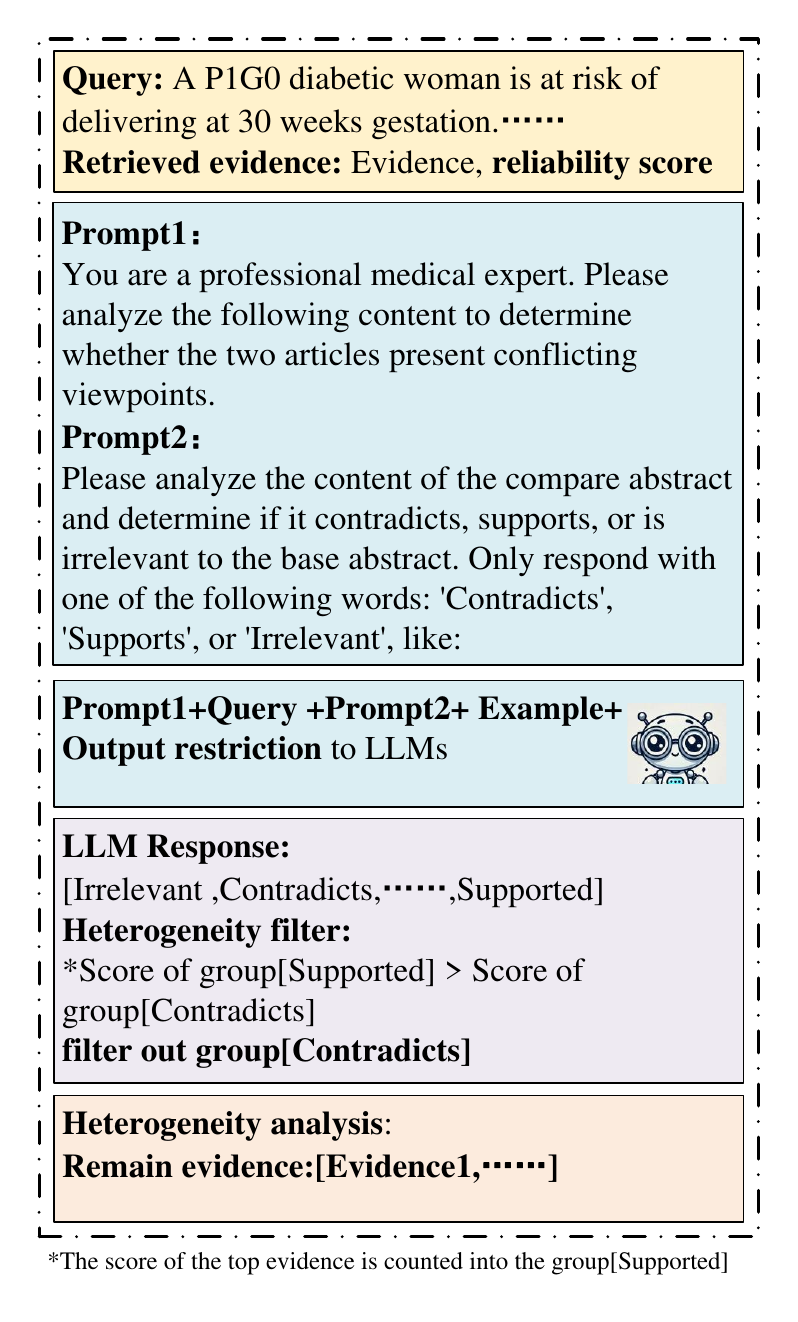}
    \caption{Details of the Heterogeneity analysis. This ensures the consistency of the evidence provided to the generation model. }
    \label{fig:6}
\end{figure}

Our filter method starts with selecting the highest-scoring documents from the reliability analysis as the base articles. We compare the remaining articles to this article to determine if there are supportive, irrelevant, or contradictory relationships. We employ an LLM to categorize all articles accordingly. Ultimately, we compare the total scores of articles that support the baseline against those that oppose it. The lower-scoring group has all its articles removed. This method helps retain a consistent and coherent set of evidence that enhances the quality and relevance of the information fed into the generation model.

\subsection{Extrapolatation Analysis }
\begin{figure}[h]
    \centering
    \includegraphics[width=1.0\linewidth]{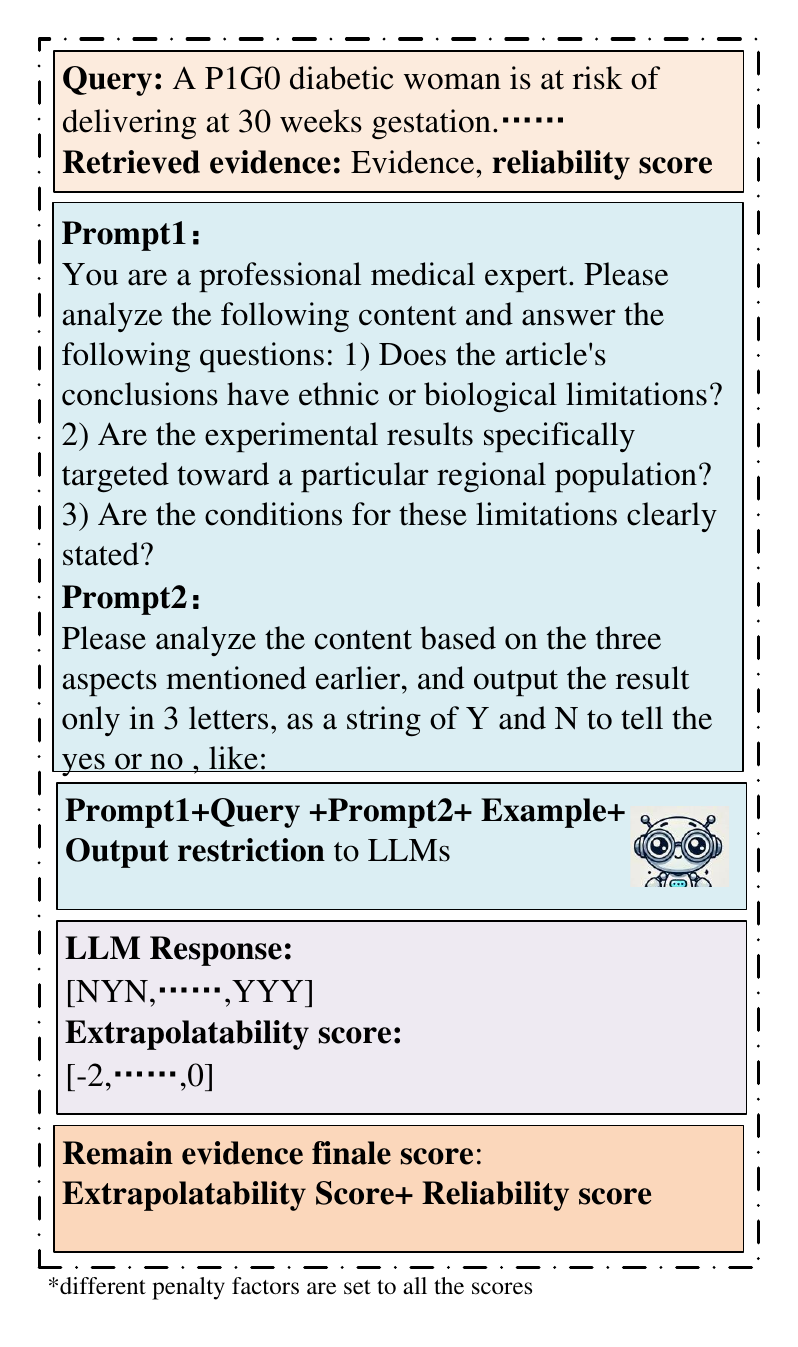}
    \caption{Details of the Extrapolation analysis. This adjustment ensures that we obtain a score that truly represents the quality of the help to the users.}
    \label{fig:7}
\end{figure}

Finally, to prevent a significant discrepancy between the user's biological condition and the experimental results, we need to adjust the scores based on their extrapolatability. For the articles after the heterogeneity analysis, we prompt the LLM to discern whether the conclusions of these articles have any racial or biological limitations and whether the experimental results could cause significant side effects for specific populations as shown in Figure \ref{fig:7}. We adjust the scores positively for articles without conditions, negatively for articles with applicable conditions, and neutrally for those with unclear conditions.
This adjustment ensures that we obtain a score that truly represents the quality of the help to the users. After this analysis, we can provide the top-scored evidence to the generation model, resulting in a safe and reliable high-quality response. This process not only enhances the applicability of the responses but also ensures they are tailored to the specific biological context of the user, increasing the accuracy and safety of the medical advice provided.

\begin{figure*}[h]
    \centering
    \includegraphics[width=0.8\linewidth]{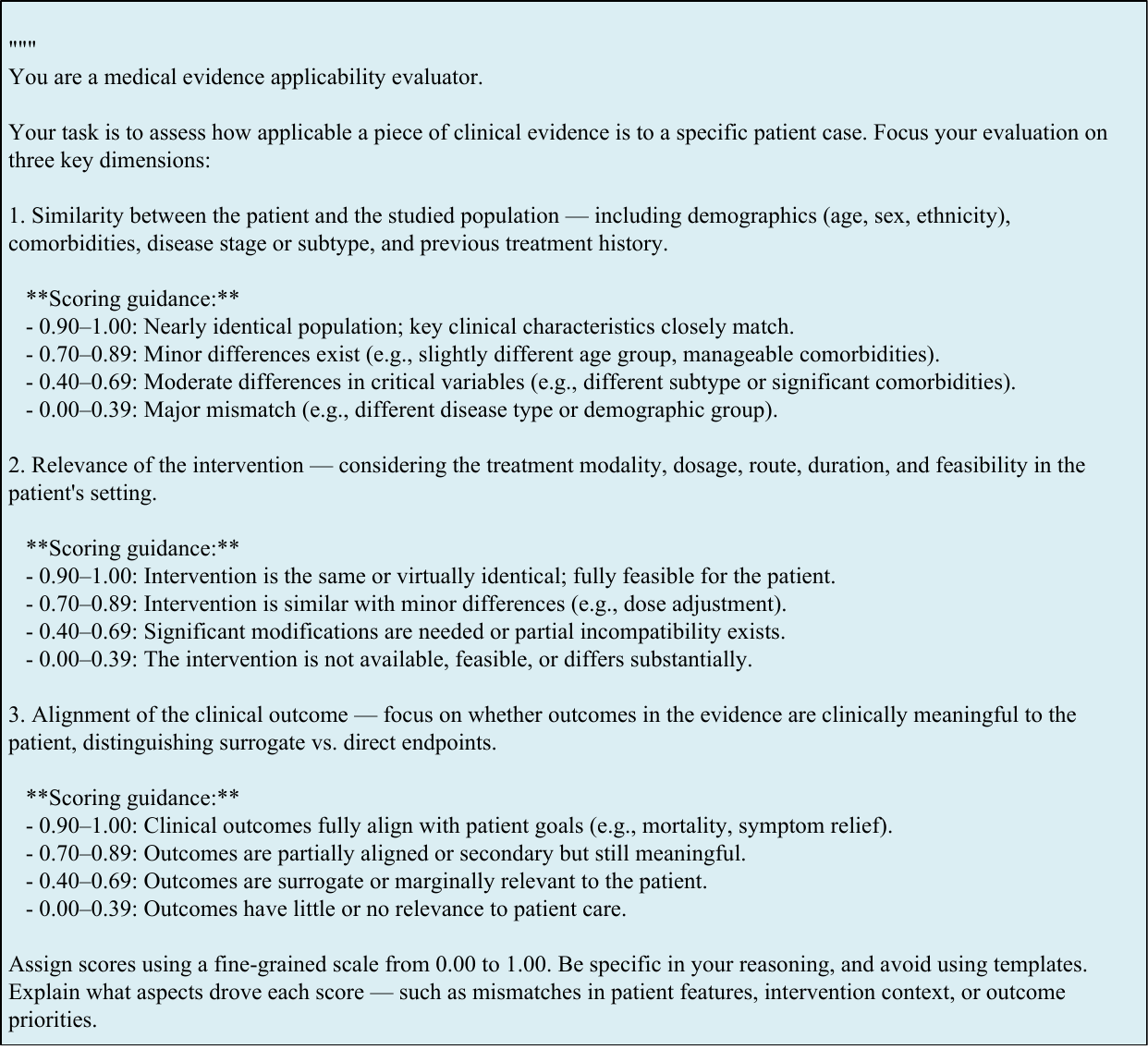}
    \caption{Details of the Extrapolation analysis prompts. }
    \label{fig:8}
\end{figure*}

However, in our empirical experiments, we find that the large model’s extrapolability assessment was overly coarse-grained. As a result, it tend to judge the vast majority of evidence as non‑extrapolable, rendering the evaluation meaningless. To address this, we adapt the prompt shown in Figure~\ref{fig:8}, which provides a more fine‑grained criterion for directly assessing the extrapolability of each piece of evidence to the user’s background.

\subsection{Experiments}

As shown in Table \ref{tab:2}, We employ another evaluation metric to evaluate the evidence quality. We assess the contribution of our input evidence to the evidence options as a measure of evidence quality. We observe that after reranking and filtering, the average quality of the evidence is effectively enhanced. Additionally, We also analyze the proportion of articles that have positive evidence relevance. Our method successfully demonstrates that good evidence has been provided for the generation model. The evidence extracted by META-RAG is more closely aligned with the correct answers among the options.
\begin{table}[htbp]

\begin{center}
\begin{tabular}{c c c}\hline\hline
\textbf{Methods}&\textbf{*ECS}&\textbf{*PPA}\\
\hline
\textbf{Baseline1}  & - & -  \\
\textbf{Baseline2} & -1.8924 & 12.33\% \\
\textbf{Baseline3} & -0.8355 & 12.33\% \\
\textbf{META-RAG} & -0.5718 & 12.62\% \\
\hline
\hline
\end{tabular}
\caption{Evaluation of the evidence retrieved by Meta-RAG and other Baselines. The ECS is short for evidence contribution score which represents the average similarity between the articles and each option. And PAP is the proportion of the positive article. These two scores are invalid for Baseline 1 because of the unrelated evidence.}
\label{tab:2}
\end{center}
\end{table}

\begin{table}[htbp]
\begin{center}
\begin{tabular}{c c c c}
\hline
\hline
\textbf{Methods}&\textbf{Accuracy}&\textbf{ECS}&\textbf{PPA}\\
\hline
\textbf{EP=-0.2} & 46.60\%& -0.5718 & 12.62\%\\
\textbf{EP=-0.3} & 41.75\%& -0.7156 & 14.56\%\\
\textbf{EP=-0.4} & 37.86\%& -0.5800 & 12.62\%\\
\hline
\hline
\end{tabular}
\caption{Ablation of Extrapolatability penalty.}
\label{tab:4}
\end{center}
\end{table}
Also, we test the extrapolatability analysis component of the model. The results mainly reflect whether the evidence can effectively serve the user. While this issue occurs less frequently in medical problems, it is critical in EBM. We include this module to ensure that users can access relevant knowledge more precisely. For LLMs, the influence of this part seems less than the other two. Therefore, as shown in Table~\ref {tab:4}, we implement a more detailed experiment to test the influence of extrapolatation penalty score. The experimental results show that the model's accuracy is highly sensitive to the penalty coefficient. With an incorrect penalty coefficient, the output accuracy significantly decreases, which indirectly demonstrates the necessity of this step in the process.

\subsection{How about other LLMs}
We also evaluated several of today’s top-performing large language models. As illustrated in Figure~\ref{fig:6}, we report GPT‑4o’s performance on the two datasets used in our study. We observe substantial instability across different baselines. On the MEDQA dataset, GPT‑4o achieves its best results when answering directly—without any supporting evidence—whereas on the MMLU dataset, supplying random evidence yields superior performance.

For MEDQA, we hypothesize that GPT‑4o was trained on an excessive amount of the same or very similar data, leading it to rely far more heavily on memorized internal evidence than on external inputs. As a result, providing additional evidence actually misleads the model and degrades its performance. By contrast, for MMLU, GPT‑4o appears not to have been exposed to equivalent training data; consequently, the direct-answering baseline produces the worst results. In this case, the injected evidence does not constrain the model’s reasoning path but rather “awakens” its latent knowledge, enabling it to recall relevant facts more clearly.

Taken together, these findings suggest that for widely studied benchmarks like MEDQA and MMLU, large models already possess substantial internal reserves of knowledge. As a result, applying our method to such over‑provisioned models yields only marginal gains. To fully demonstrate the advantages of our approach, it will be necessary to evaluate on more challenging datasets with less pre‑exposure.
\begin{figure}[h]
    \centering
    \includegraphics[width=1.0\linewidth]{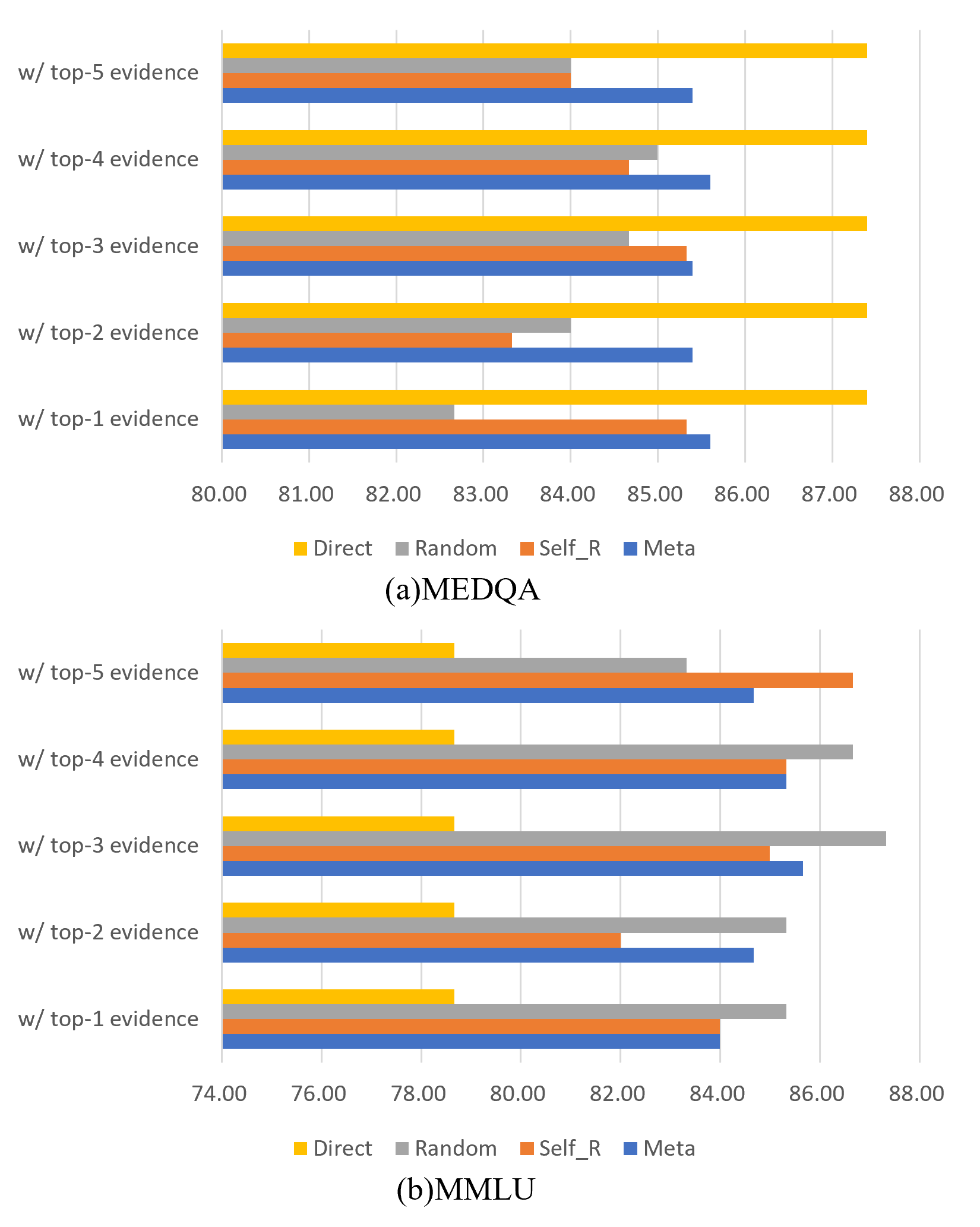}
    \caption{Performance curves of the GPT‑4o model under our method. }
    \label{fig:6}
\end{figure}

\section*{Discussion}
(1)Traditional QA tasks cannot effectively evaluate our evidence ranking method. Our approach aims to guide the model to adhere to the principles of evidence screening and ranking in EBM, retrieving more effective and reliable evidence. Our method can enhance the acceptance of the screened evidence among the medical community. However, for traditional QA tasks, the improvement provided by our method is indeed not significant. This phenomenon compels us to add the level of evidence as an evaluation metric. In our subsequent work, we will seek evaluation methods that better reflect the unique characteristics of medical evidence.

(2) We cannot fully replicate every detail of the meticulous process involved in a meta-analysis. To understand the relevant principles of meta-analysis, we consulted several medical experts. We find that the heterogeneity analysis step typically requires detailed experimental data from each medical paper to perform calculations and comparisons. However, our LLM cannot email each author to obtain this data. Therefore, our heterogeneity analysis is merely a coarse-grained exclusion method based on statistical approaches. In our future work, we will attempt to refine this process to make it more precise and credible.

(3)Our extrapolation method does not significantly influence the final judgment formed by the generation model. We can only enhance its utility as a reference by assigning increased weights. However, in the medical evidence filtering process, evidence with excessive limitations cannot be applied to users. Unfortunately, the LLMs are hard to make the decision only based on literature abstracts. Therefore, we have to implement this component from other perspectives as a scoring system based on the number of limitations. In our future work, we will optimize this extrapolation evaluation system to make it more reasonable.
\section*{Ethical Statement}
All medical data are public and open-source with no information leakage.